%% file: main.tex
\documentclass[runningheads]{llncs}

\usepackage{eccv}

\usepackage{eccvabbrv}

\usepackage{graphicx}
\usepackage{booktabs}

\usepackage[accsupp]{axessibility}  % Improves PDF readability for those with disabilities.

\usepackage{hyperref}

\usepackage{orcidlink}

\usepackage{orcidlink}
\usepackage{graphicx}
\usepackage{amssymb}
\usepackage[figuresright]{rotating}
\usepackage{amsmath}
\usepackage{booktabs}
\usepackage{algorithm}
\usepackage{algorithmicx}
\usepackage{algpseudocode} 
\usepackage{booktabs}
\usepackage{url}
\usepackage{multirow}

\usepackage{color}
\usepackage{soul}
 
\usepackage{changes}
\usepackage[capitalize]{cleveref}
\crefname{section}{Sec.}{Secs.}
\Crefname{section}{Section}{Sections}
\Crefname{table}{Table}{Tables}
\crefname{table}{Tab.}{Tabs.}
\definecolor{mistyrose}{rgb}{1.0, 0.89, 0.88}

\newcommand{\net}{CLAST\ }

\begin{document}

\title{Bridging Text and Image for Artist Style Transfer via Contrastive Learning} 

\titlerunning{Abbreviated paper title}

\author{Zhi-Song Liu\inst{1}\orcidlink{0000-0003-4507-3097} \and
Li-Wen Wang\inst{2}\orcidlink{0000-0001-9548-7795} \and
Jun Xiao\inst{2}\orcidlink{0000-0002-4935-7866} \and
Vicky Kalogeiton\inst{3}\orcidlink{0000-0002-7368-6993}
}

\authorrunning{Z.~Liu et al.}

\institute{Lappeenranta-Lahti University of Technology LUT, Finland \\
\email{zhisong.liu@lut.fi} \and
The Hong Kong Polytechnic University, Hong Kong \and
LIX, Ecole Polytechnique, IP Paris
}

\maketitle

\begin{abstract}
Image style transfer has attracted widespread attention in the past few years. Despite its remarkable results, it requires additional style images available as references, making it less flexible and inconvenient. Using text is the most natural way to describe the style. More importantly, text can describe implicit abstract styles, like styles of specific artists or art movements. In this paper, we propose a Contrastive Learning for Artistic Style Transfer (CLAST) that leverages advanced image-text encoders to control arbitrary style transfer. We introduce a supervised contrastive training strategy to effectively extract style descriptions from the image-text model (i.e., CLIP), which aligns stylization with the text description. To this end, we also propose a novel and efficient adaLN based state space models that explore style-content fusion. Finally, we achieve a text-driven image style transfer. Extensive experiments demonstrate that our approach outperforms the state-of-the-art methods in artistic style transfer. More importantly, it does not require online fine-tuning and can render a $512\times 512$ image in 0.03s.
\end{abstract}

\input{Section/01_Introduction}

\input{Section/02_Related_Works}
\input{Section/04_Approach}
\input{Section/05_Experiments}
\input{Section/06_Conclusion}

\section*{Acknowledgements}
This work was supported by ANR-22-CE23-0007, Hi!Paris collaborative project and DIM RFSI 2021.

\bibliographystyle{splncs04}
\bibliography{main}
\end{document}

%% file: Section/01_Introduction.tex
\section{Introduction}
\label{sec:introduction}

Image style transfer is a popular topic that aims to apply the desired painting style to an input content image. The transfer model requires the information of \textit{``what content"} in the input image and \textit{``which painting style"} to be used~\cite{ast,language_2}. Conventional style transfer methods require a content image accompanied by a style image to provide the content and style information ~\cite{adain,adaattn,sanet,art-net,artflow,cast,stytr2,quantart}. However, people have specific aesthetic needs. Usually,  finding a single style image that perfectly matches one's requirements is inconvenient or infeasible. Text or language is a natural interface to describe the preferred style. Instead of using a style image, using text to describe style preference is easier to obtain and more adjustable. Furthermore, achieving perceptually pleasing artist-aware stylization typically requires learning from collections of art, as one reference image is not representative enough. 
\begin{figure}[t]
	\centering
		\centerline{\includegraphics[width=\columnwidth]{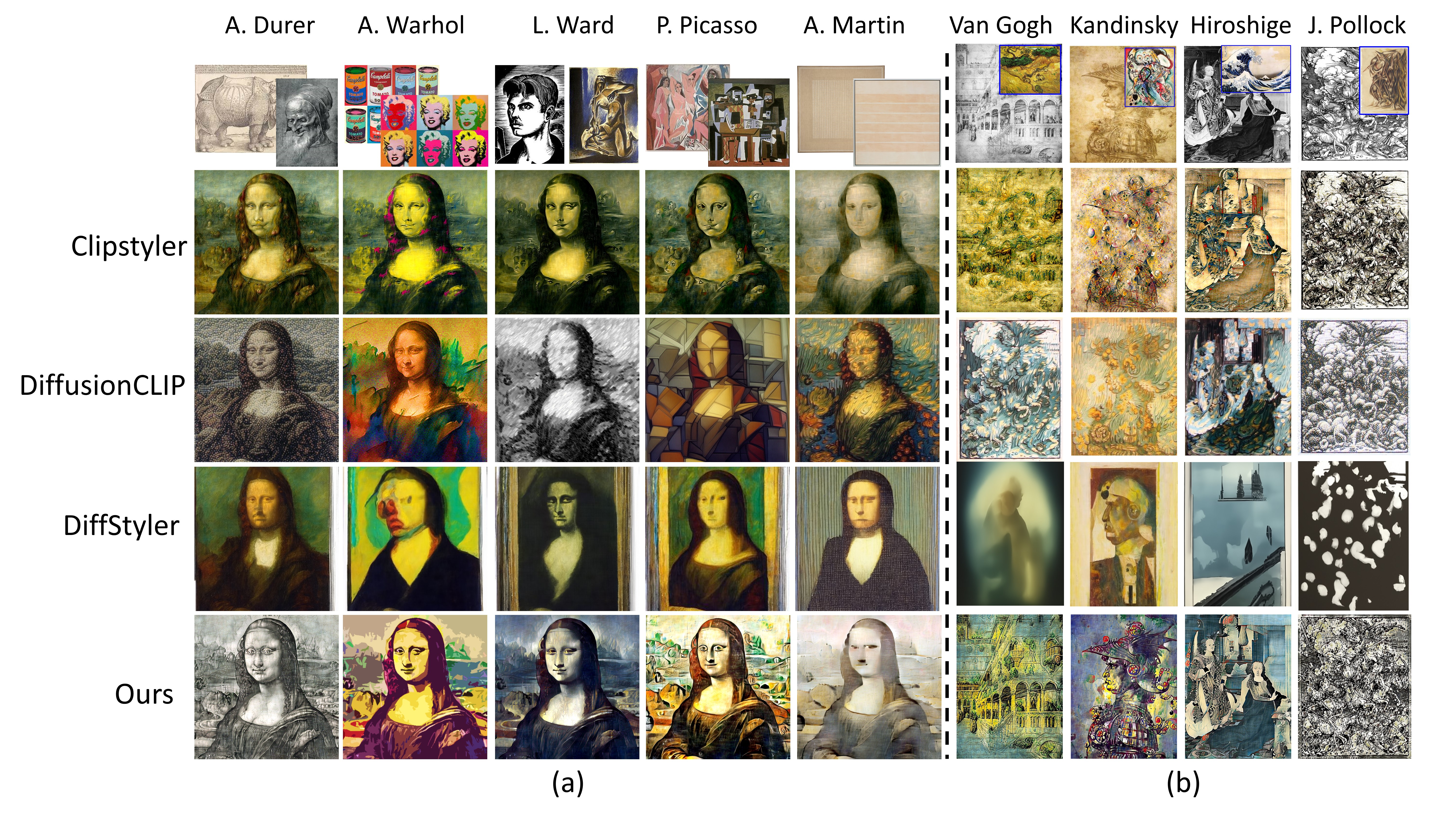}}
		\caption{\small{\textbf{Artist-aware style transfer.} For (a), given five artists (top row) and the text prompt \textit{``a Mona Lisa painting in Pablo Picasso style''}, we replace the artist’s name in the prompt with each of the five and get stylization results. We show results for Clipstyler~\cite{language_2}, DiffusionCLIP~\cite{diffusionclip}, DiffStyler~\cite{diffstyler}, and our proposed \net. Our method mimics the most representative styles from each artist, while Clipstyler~\cite{language_2} shows little style changes, DiffStyler changes facial features, and DiffusionCLIP does not learn representative styles. For (b), we show the stylization of four sketch paintings (top row). We can see that ours can transfer the styles without affecting the contents.
		}}
		\label{fig:teaser}
\end{figure}

In this work, we learn arbitrary artist-aware image style transfer, which transfers the painting styles of any artist to the target image using just texts. Most studies on universal style transfer~\cite{LT,sanet,ast} limit their applications to using reference images as style indicators that are less creative or flexible. Text-driven style transfer has been studied~\cite{language_1,language_2} and has shown promising results using a simple text prompt. However, these approaches require either costly data collection and labeling or online optimization for every content and style (e.g. DiffusionCLIP~\cite{diffusionclip} and Clipstyler~\cite{language_2} in Figure~\ref{fig:teaser}). 
Instead, our proposed Text-driven artistic aware Style Transfer model \net overcomes these two problems and achieves better and more efficient stylization.  
Figure~\ref{fig:teaser} illustrates this. It shows the artist-aware stylization on \textit{Mona Lisa} by comparing text-driven methods (Clipstyler~\cite{language_2}, DiffusionCLIP~\cite{diffusionclip}, DiffStyler~\cite{diffstyler} and Ours) on five artists. We input text prompts such as \textit{``a Mona Lisa painting in Pablo Picasso style''} and change the artist's name accordingly. \net faithfully mimics the painters' style and preserves the contents reliably. Figure~\ref{fig:teaser} (b) shows four examples of transferring four different sketches to the target styles, we can see that ours can better preserve the contours and lines of the original sketches, while others fail to achieve this.

%To obtain artist awareness, our proposed \net explicitly explores the hidden space where the global distance of different artworks (different artists) can be maximized, while the same artworks (same artists) can be minimized. It can resonate with the classic Fisher Linear Discriminant~\cite{fda} to distribute features into groups. Specifically, given artists' names, we project features from different artists onto the CLIP space for classification. 
%
%To prove the text-image correlations, we empirically analyze the co-linearity between artists\footnote{We choose artists' names rather than art movements (e.g. impressionism or cubism) is that each artist has relatively consistent painting styles while one art movement can includes complete different painters. For example, Paul Cezanne and Claude Monet are both impressionists. Cezanne's painting is more colorful and imaginative while Monet's painting is clear and natural.~\cite{Monet_and_Cezanne}} and paintings on the CLIP space and demonstrate the reasonableness and effectiveness of text-driven style transfer. Our experiments on both text- and image-driven style transfer show that \net outperforms the previous state of the art quantitatively and qualitatively. 
To obtain artist awareness, \net explicitly explores the latent space: it maximizes the global distance amongst different artworks of different artists, while it minimizes the distance amongst artworks of the same artists. 
Specifically,  artists have their style visualized in many paintings. The name of the artist is the only ``label'' that can connect artists to their painting style. \net not only is driven by different artists' names, but it also learns to group the painters' works to extract the most representative features. Specifically, given artists' names, \net projects features from different artists onto the CLIP space for classification. Moreover, we explore text-image correlations by empirically analyzing the co-linearity between artists\footnote{We use artists' names rather than art movements (e.g. impressionism or cubism) as each artist has a relatively consistent painting style while one art movement can include different painters. For instance, Paul Cezanne and Claude Monet are both impressionists. Cezanne's paintings are more colorful and imaginative while Monet's paintings are clear and natural~\cite{Monet_and_Cezanne}.} and paintings in the CLIP space demonstrate the effectiveness of text-driven style transfer. To achieve real-time inference, we adopt the adaLN based State Space Model (adaLN-SSM) to realize the style fusion as a linear sequential regression. It can greatly reduce the computation time and improve the style transformation. Our contributions can be summarized as follows:
(1) To achieve text-driven image style transfer, we propose to embed the task-agnostic CLIP image-text model into our proposed \net. This enables \net to obtain style preference from \textit{text descriptions}, making the image style transfer more interactive. 
(2) We propose a adaLN-based state space model (adaLN-SSM) to explore style-content fusion, which can efficiently model both local and global feature correlations. The stylized image not only can visualize the statistical similarity to the text description, but also preserve the original contents.
(3) We suggest using a supervised contrastive training strategy (Sections~\ref{sub:txst}) to learn art collection awareness. It can align corresponding artistic texts and images offline so the model can apply stylization in real time.
(4) We conduct extensive experiments on text-driven style transfer. We show that \net outperforms the state-of-the-art methods both quantitatively and qualitatively.

%% file: Section/02_Related_Works.tex
\section{Related Work}
\label{sec:relwork}
\noindent \textbf{Arbitrary style transfer.} It can be split into two groups: (1) style-aware optimization~\cite{Gatys,Ulyanov,wct,strotss} and (2) universal style transfer~\cite{adain,LT,avatar,art-net}. 
%The former explicitly studies the statistic correlations between content and style images in the deep feature space. Gatys et al.~\cite{Gatys} propose the first flexible iterative optimization approach based on a pre-trained VGG19 network~\cite{VGG}. It can achieve arbitrary style transfer, but the forward and backward passes are time-consuming. STROTSS~\cite{strotss} replaces the l-1 style loss to Relaxed EMD distance to achieve higher visual quality. 
%WCT~\cite{wct} studies the second-order correlations via whitening and coloring transforms without costly optimization. To further reduce the computation of WCT, LT~\cite{LT} uses linear transformations to approximate the matrix decomposition, resulting in real-time style transfer. 
Inspired by the attention mechanism~\cite{attention}, a few works use it to explore statistical correlations. SANet~\cite{sanet} matches the content and style statistics via cross-attention. AdaAttN~\cite{adaattn} further explores the second-order attention to preserve more content information without losing style patterns. Artflow~\cite{artflow} and VAEST~\cite{vaest} investigate the normalization flow~\cite{flow} and VAE~\cite{vae} to fuse style and content images. The latter focuses on zero- or first-order statistics for real-time style transfer. AdaIN~\cite{adain} proposes Adaptive instance normalization to shift the deep content feature to the style space. %Avatar-net~\cite{avatar} proposes to combine AdaIN~\cite{adain} and style swap~\cite{swap} to enhance the stylization in a coarse-to-fine manner.
ReReVST~\cite{rerevst} follows Avatar-net~\cite{avatar} to further develop inter-channel feature adjustment for both image and video style transfer. ArtNet~\cite{art-net} proposes Zero-channel Pruning to reduce model complexity. More recently, there are some developments in transformer-based style transfer~\cite{stytr2,vision_transformer_2} that use the vision transformer~\cite{vit} for stylization. \cite{all_to_key} modify the attention modules to pay attention to the fine-grained styles. Most recently, \cite{quantart} proposes vector quantization to achieve latent space feature fusion for stylization. \cite{inverter,stylediffusion} propose to utilize the power of a pre-trained diffusion model for image-based style transfer.

\noindent\textbf{Artistic style transfer.} Arbitrary-style transfer methods suffer from the fact that they learn style statistics from one reference image, which is not representative enough of the desired pattern. Artistic style transfer learns robust statistics from many art works so that it can transfer the most distinctive styles to the content images. A straightforward approach is to collect a number of desirable images to train a specific model for style transfer~\cite{dufour2022scam}. For example, AST~\cite{ast} proposes an artist-aware style transfer to achieve art stylization. In other words, they collect specific artworks as style references, e.g., Van Gogh, to train a network for specific style transfer. DualAST~\cite{dualast} follows this idea and proposes a more flexible network to balance both artwork style and artist style via Style-Control Block. CAST~\cite{cast} learns the style similarities and differences between multiple styles directly from image features. MAST~\cite{mast} proposes a per-pixel process via style kernel, which can dynamically adapt to the contents for artistic style transfer. 

% \begin{figure*}[t]
% \centering
% 		\centerline{\includegraphics[width=\linewidth]{Figure/method_compare.jpg}}
% 		\caption{\small{\textbf{Comparison among different CLIP based image manipulation.} We show the overall training processes of a) StyleCLIP, b) Clipstyler and c) our proposed \net for comparison. They all use CLIP image encoder and text encoder to project images and texts onto the CLIP space, obtaining corresponding features $E_I$ and $E_T$, for distance $\Delta T$ measurement. For a), it samples latent code $\omega$ from the pre-define latent space $\mathcal{W}+$ as input to StyleGAN for a generation. For b), it uses source image $I_c$ as input to the StyleNet to obtain $f(I_c)$. For c), it randomly samples different style images from different artists (e.g., Van Gogh and Picasso) as, $I_s(1,j)$ and $I_s(2,i)$. 1 and 2 stand for two artists. i and j stand for the index of different paintings. Then it applies both text-guided and image-guided style transfer. It minimizes the intra-class distances (purple arrows) $\Delta T_1$ and $\Delta T_2$, as well as maximizes the inter-class distance (yellow arrows) $\Delta D$.
% 		}}
% 		\label{fig:method_compare}
% \end{figure*}

\noindent\textbf{Text-driven image style transfer.}
Style transfer is a subjective topic, that is, different people may have different preferences for stylization.
Using style images as references may not be as sufficiently good as texts can describe styles in a more abstract and aesthetic manner.
The success of CLIP~\cite{clip}, VQVAE~\cite{vqvae} and multimodality~\cite{multi_1,multi_2,multi_3} show that text and image can be related via a shared projection space. Some pioneer works on image editing~\cite{dalle,stylegannada,clip_draw_1,clip_draw_2,styleclip,liu2021control,VisualConceptVocabulary,ldm} and video understanding~\cite{univl,clip2video,clip4clip} show that language can be used for user-guided applications. \cite{VisualConceptVocabulary} constructs open-ended vocabularies to flexibly recompose the visual content in the latent spaces of GANs. Most recently, \cite{language_1} uses the CLIP as the condition for style transfer that increases the cross-correlation between the output and text description for text-guided style transfer. Clipstyler~\cite{language_2} further developed this idea by using both global and patch CLIP losses to generate high-resolution stylized images. On a different trend, there are several approaches~\cite{language_5,language_6,language_7,language_8} that use text and/or image prompts for image synthesis and creative painting. For instance, ~\cite{language_4,lanit,mgad,language_8} use text prompts to create a novel painting by semantic editing, like face and fashion design. Their main limitation is that they synthesize the new image much closer to the text description and the contents get distorted and do not match the original content images. Most recently, diffusion models have gained great attention for text-driven style transfer. DiffusionCLIP~\cite{diffusionclip} and InST~\cite{inverter} fine-tune the pre-trained diffusion model for image manipulation, and Zecon~\cite{zecon} uses zero-shot learning to transfer patch-based patterns to the content images. StylerDALLE~\cite{styledalle} proposes to combine Dall-E~\cite{dalle} and CLIP to achieve arbitrary text-driven stylization. However, it requires hours of online optimization for one style, which makes it less applicable to practice. For this, TxST~\cite{txst,txst2} proposes to learn the artistic styles through Contrastive learning offline so that arbitrary stylization can be achieved online. Diffstyler~\cite{diffstyler} utilizes the diffusion model to take the text as the condition for cross-model guidance.

\begin{figure}[!t]
\centering
\includegraphics[width=\linewidth]{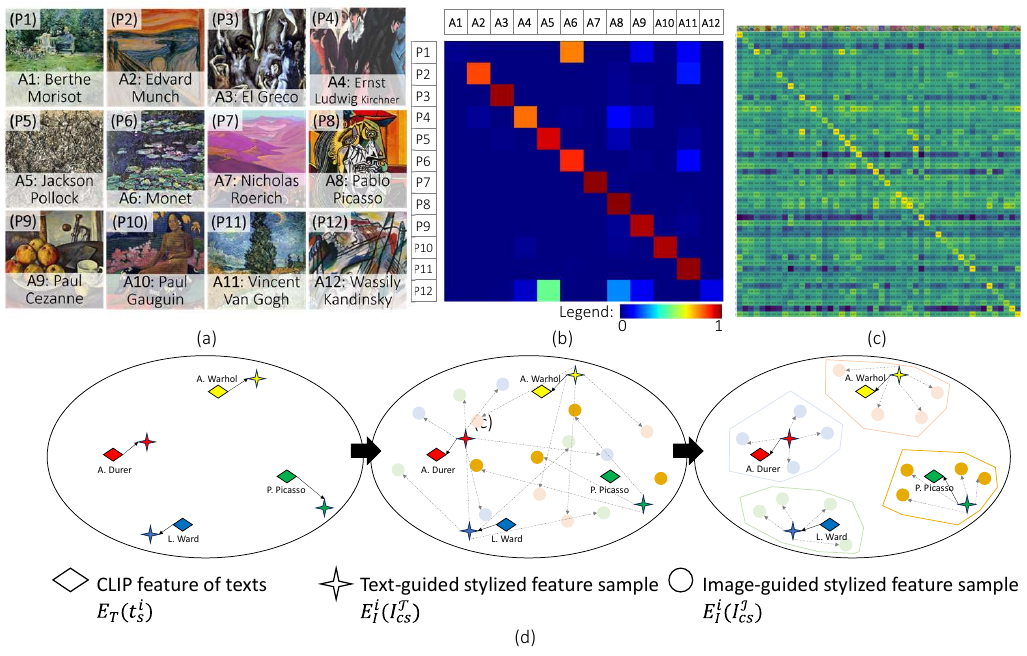}
\caption{\small{\textbf{Correlation between artists and paintings.}(a) A set of Artist-Painting paired samples from WikiArt dataset \cite{WikiArt}. The artists (Abbr.: \textbf{A}) are represented using text, and their paintings (Abbr.: \textbf{P}) are in the format of color images. Different artists and their paints are given different index numbers for clear visualization. (b) Feature relationship between artists and paintings. Features of the artists and paintings are extracted by the CLIP\cite{clip} with language and visual portions. The horizontal and vertical axes are the artists and paintings respectively. The significantly larger values on the diagonal elements suggest that the features from the CLIP model are aware of the high-level painting style of a different artist. (c) shows the painting-artist correlations on the whole WikiArt (d) shows how contrastive learning is used for style transfer.}}
\label{fig:ana_wiki}
\end{figure}

%% file: Section/04_Approach.tex
\section{Approach}
%Language for style transfer (CLIP observation, with figures) Artistic style transfer (CLIP+ST, contrastive!!!). Training, Testing

\subsection{Text for Style Transfer}
\label{sub:textstyle}
Painting style, or \textit{painting language}, represents the painting tastes of artists. It is a high-level abstract representation of the images. Previous style transfer methods~\cite{adain,art-net,sanet} use a referenced image to describe the desired painting style. The core idea of our approach is to describe painting style using texts, because of its convenience and flexibility. The benefits of the text model are twofold: (1) high-level style representation that avoids ambiguity caused by the image content; and (2) flexible adjustment with no effort for searching proper references. Recently developed text models, like CLIP\cite{clip}, motivate us to ask \textit{``Can text models be aware of different style representations?"} 

We believe that text and image can work interchangeably in the CLIP space for style transfer. In other words, text and image are co-linear in the CLIP space and hence, they can both be used as style indicators. From the aspect of the style description, this co-linearity also exists between artists and their paintings, making it possible to realize artist-aware style transfer. To verify this, we have made an analysis between images and style queries using a pre-trained CLIP model~\cite{clip}. Initially, we collected a set of Artist-Painting pairs from WikiArt dataset~\cite{WikiArt}, as shown in Figure~\ref{fig:ana_wiki}(a). To investigate the correlation between artists and paintings, the names of the artist are first tokenized and encoded to extract the feature $E_{T}\in\mathbb{R}^{512}$. Next, we encoded the paintings for the image feature $E_{I}\in\mathbb{R}^{512}$. The correlation was then calculated by the dot product between the text $E_{T}(t_s)$ and image features $E_{I}(I_s)$. For each painting $I_s^i$, we used the softmax function to find the probability score $s$ for different artists $t_s^j$, as follows:
\begin{small}
\begin{equation} \tag*{(1)}
\label{eq_ana_clip}
s\left( i,j \right) =\frac{\exp \big(S\left( E_{I}(I_s^i)\cdot E_{T}(t_s^j) \right)\big)}{\sum\nolimits_{k=0}^{k=C}{\exp \big(S\left( E_{I}(I_s^i)\cdot E_{T}(t_s^k) \right)\big)}}
\end{equation}
\end{small}

\noindent where $S$ is the cosine function, $C$ is the number of artists. A larger score $s\left( i,j \right)$ means more likely the painting $I_s^i$ is drawn by the artist $t_s^j$. The relationship among scores is shown in Figure~\ref{fig:ana_wiki}(b), where we observe the significantly larger values of the diagonal. Intuitively, we observe a strong correlation between the artists and their paintings in the CLIP feature space.\footnote{There are two outliers in the figure, i.e., ``\textbf{P}1-\textbf{A}6" and ``\textbf{P}12-\textbf{A}5". They indicate that the painting \textbf{P}1 by \textit{Berthe Morisot} has the highest score to the painting of artist \textit{Monet}(\textbf{A}6). The reason is that both artists lived in the same country (France), and the artworks were created during the same artistic period (both were born in the 1840s) which we call the Impressionism movement. For another outlier ``\textbf{P}12-\textbf{A}5", the painting from \textit{Wassily Kandinsky} has similar styles to \textit{Jackson Pollock}. Both paintings \textbf{P}5 and \textbf{P}12 in Figure~\ref{fig:ana_wiki}(a) use geometric symbols in complex and abstract forms that are difficult to distinguish.} The same painting-artist correlations are also observed from the complete WikiArt (Figure~\ref{fig:ana_wiki}(c)).

\subsection{Proposed text-driven image style transfer (\net)}
\label{sub:txst}
\begin{figure}[t]
	\begin{center}
		\centerline{\includegraphics[width=\columnwidth]{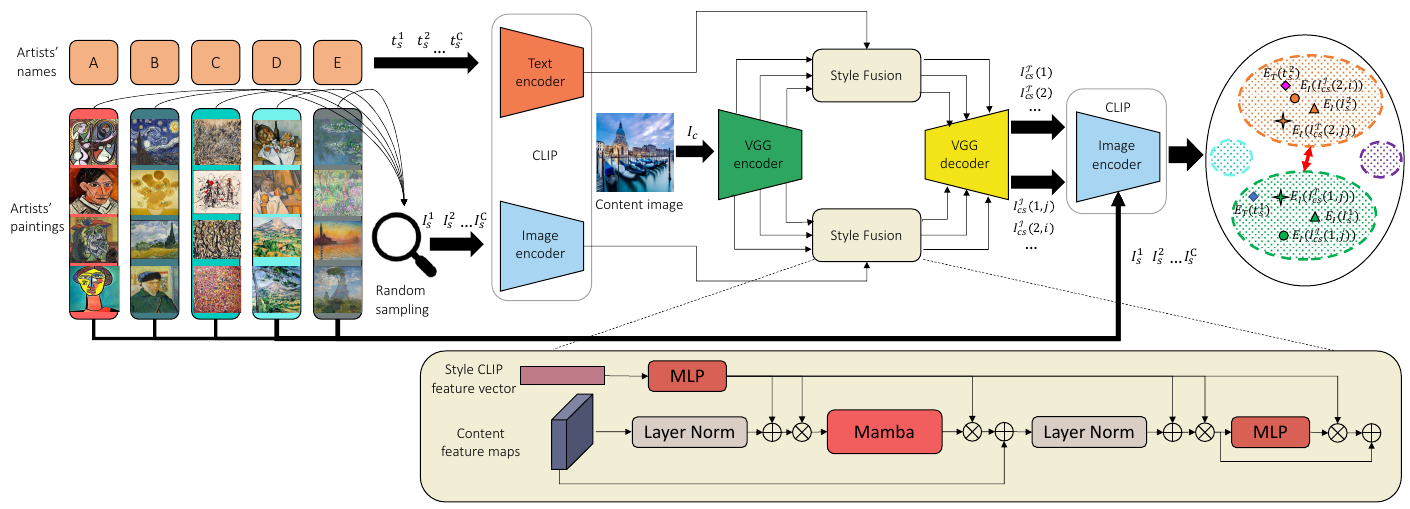}}
		\caption{\small{\textbf{Training process of the proposed \net.} We show the overall training processes of our proposed \net. Given content images, artists' names and paintings, we process them by CLIP to cluster stylized results generated by texts or images. The Style Fusion module utilizes the Adaptive layer norm and Mamba structure to transfer textual or visual styles to the content images.
		}}
		\label{fig:network}
	\end{center}
\end{figure}

Figure~\ref{fig:ana_wiki}(d) highlights the supervised contrastive learning process about how we can use CLIP to achieve text-driven style transfer. The CLIP loss helps us find the co-linearity between style texts and output images (the solid lines between diamonds and stars), while the contrastive similarity loss minimizes the feature distance between text-guided results $E^i_I(I^\mathcal{T}_{cs})$ and image-guided results $E^i_I(I^\mathcal{I}_{cs})$ (dashed lines between stars and circles). The names of artists ($t_s$) also work as labels to supervise the contrastive training process. Therefore, we guide the model to minimize the intra-class distance between different art collections painted by the same painter, as well as maximize the inter-class distance between different painters. Based on this, in this section, we introduce the proposed \net for artist-aware image style transfer, which is illustrated in Figure~\ref{fig:network}.

Given content images $I_c$ and desirable styles, artist's name $t_s$ and corresponding painting $I_s$, \net outputs the stylized image $I_{cs}$. We denote text-guided stylized images as $I^\mathcal{T}_{cs}$ and image-guided stylized images as $I^\mathcal{I}_{cs}$. In the training stage, we first use CLIP to project images to the latent space. To preserve style consistency, the network computes the contrastive similarity to minimize the inter-distance of paired stylized images. We propose \textbf{Supervised Contrative loss (SupCon loss)} to ensure that given one artist's name, the generated $I^\mathcal{T}_{cs}$, $I^\mathcal{I}_{cs}$ and target paintings $I_s$ can be clustered together with uniformly distinct feature representation. At test time, either text ($t_s$) and image ($I_s$) can be input as style indicators to generate consistent stylization.

\noindent \textbf{Architecture.} 
\net consists of four parts: Image encoder, Image decoder, CLIP (text and image encoders), and Style fusion (adaLN based state space model). The structure of the image encoder follows VGG-19~\cite{VGG}, discarding the fully connected layers. The image decoder is symmetric to the image encoder, with gradual upsampling feature maps towards final stylized images. We can use CLIP to encode the paired texts ($t_s^i, i=1,2,..,C$, where $C$ is the number of artists) and images ($I_s^i, i=1,2,..,N$, where $N$ is the number of paintings) to obtain corresponding text features $E_\mathcal{T}(t_s^i)$ and image features $E_\mathcal{I}(I_s^i)$, respectively. The VGG encoder encodes the content image, and then the Style Fusion module fuses the style and content features. Finally, the VGG decoder outputs the stylized results. Below, we introduce the proposed Style Fusion module in detail.

\noindent $\bullet$ \textbf{Style Fusion.} Recently, DiT~\cite{dit} has shown that the Adaptive Layer Norm (adaLN) works well on the conditional diffusion model. The idea is to regress the conditional vector to obtain dimension-wise scaling and shift parameters and use them to adjust the target distribution. It resonates with the AdaIN~\cite{adain} process that we can use adaptive normalization to shift the content features to the style domain for style manipulation. Following this idea, we incorporate adaLN into Style Fusion. As can be seen from Figure~\ref{fig:network}, the CLIP feature vector $z$ is regressed by MLP to output scale and shift parameters that can represent feature distributions. The content feature $x$ is renormalized by the learned style features as,
\begin{small}
\begin{equation} \tag*{(2)}
y1=(LN(x + \alpha_1 \cdot SSM(LN(x)\cdot \mu_1 + \sigma_1)) + \alpha_2) + \sigma_2,\ \ y = y1 + \alpha_2 \cdot MLP(y1)
\label{eq:adaln} 
\end{equation}
\end{small}

\noindent Equation~\ref{eq:adaln} illustrates that style fusion is a learnable normalization process. The target feature vector (generated from style texts or images) generates scale and shift vectors that can transform the content feature maps to the target style domain. $Mamba$ implements state space modelling~\cite{mamba} process, which learns global feature correlations as an efficient sequence transformation. 

\subsection{Loss functions} 
\label{sub:losses}
\noindent $\bullet$  \textbf{Directional CLIP loss.} To guide the content image following the semantic of the target text (artist's name), we use the directional CLIP loss~\cite{stylegannada,language_2} to align the CLIP-space direction between the text-image pairs of the target $t_s$ and output $I^\mathcal{T}_{cs}$. It is defined as:
\begin{small}
\begin{equation} \tag*{(3)}
\Delta T=E_T(t_s)-E_T(t_o),\ \ \Delta I=E_I(I^\mathcal{T}_{cs})-E_I(I_c),\ \ L_{clip}=1-\frac{\Delta I \cdot \Delta T}{|\Delta I| |\Delta T|}
\label{eq:clip}
\end{equation}
\end{small}

\noindent where $t_o$ is the text description for the content image. We define it as \textit{Photo} following the design in ~\cite{stylegannada}, which indicates no stylization is applied. Compared to original CLIP loss~\cite{clip}, equation~\ref{eq:clip} can stabilize the optimization process and produce results with better quality.

\noindent $\bullet$  \textbf{Supervised Contrative loss (SupCon loss).} To encourage image features to be correlated with the target style and uncorrelated with other styles, we propose to use Supervised Contrastive similarity loss (SupCon loss)~\cite{simclr} to maximize the style similarity between different paintings of the same painters. Given one content image $I_c$, and a set of $n$ target artists (name, painting, label) $\{t_s^i, I_s^i, y_i\}_{i=1}^n$, $2n$ training pairs can be created by randomly selecting either artist's name or his/her painting for style transfer, $\{\hat{t}_s^i, \hat{I}_s^i, \hat{y}_i\}_{i=1}^{2n}$. Mathematically, we have,
\begin{small}
\begin{equation} \tag*{(4)}
\label{supcon}
L_{supcon}(I^\mathcal{I}_{cs}, I^\mathcal{T}_{cs}) = -\sum_{i=1}^{2N}\frac{1}{2|N_i|-1}\sum_{j\in N(y_i),j\neq i} log\frac{exp(z_i\cdot z_j / \tau)}{\sum_{k\in I, k\neq i}exp(z_i\cdot z_k/ \tau)}
\end{equation}
\end{small}

\noindent where $\tau$ is the temperature factor, $z_k=Norm(Proj(E_I(I^\mathcal{I}_{cs}(k))))$, in which $Norm(\cdot)$ is the l2 norm operation, $Proj(\cdot)$ is a Linear layer of size 128, $E_I(\cdot)$ is the CLIP image encoder that encodes either stylized results $I^\mathcal{I}_{cs}$ edited by paintings or results $I^\mathcal{T}_{cs}$ edited by names. $N_i=\{j\in I: \hat{y}_j=\hat{y}_i \}$ contains a set of indices of samples with label $y_i$. For one specific artist, the style similarities should exist in the paintings and stylized images. Hence, we can incorporate two more pairs as $L_{supcon}(I_{s}, I^\mathcal{T}_{cs})$ and $L_{supcon}(I_{s}, I^\mathcal{I}_{cs})$. The final SupCon loss is,
\begin{small}
\begin{equation} \tag*{(5)}
\label{eq:sucon_final}
L_{supcon} = L_{supcon}(I^\mathcal{I}_{cs}, I^\mathcal{T}_{cs}) + L_{supcon}(I_{s}, I^\mathcal{T}_{cs}) + L_{supcon}(I_{s}, I^\mathcal{I}_{cs})
\end{equation}
\end{small}

\noindent $\bullet$  \textbf{Content and style feature loss.} Following existing style transfer methods~\cite{Johnson,sanet,vaest}, we employ content and style feature losses by using the pre-trained VGG network to minimize the distance in the feature space as:
\begin{small}
\begin{equation} \tag*{(6)}
L_{sty} = \sum_i^4 \Vert W_s^i(I^\mathcal{I}_{cs})\times W_s^i(I^\mathcal{I}_{cs})^T-W_s^i(I_s)\times W_s^i(I_s)^T\Vert_{1}^{1}\ \ L_{con} = \sum_i^2 \Vert W_c^i(I^\mathcal{I}_{cs})-W_c^i(I_c)\Vert_{1}^{1} 
	\label{eq:con_sty}
	\end{equation}
\end{small}

\noindent As in~\cite{Johnson}, we use VGG-19 to extract $W_s$ features (\textit{relu1\_2}, \textit{relu2\_2}, \textit{relu3\_4}, \textit{relu4\_1}) to compute the style loss. We also extract $W_c$ features (\textit{relu2\_2}, \textit{relu3\_4}) to compute the content loss. 

\noindent $\bullet$  \textbf{Total loss.} We train the network using the losses from~\cref{eq:clip,eq:sucon_final,eq:con_sty}. We also utilize LPIPS loss~\cite{LPIPS} ($L_{lpips}$) to supervise perceptual similarity. Hence, we define the final loss as $L=\lambda_{clip}L_{clip}+\lambda_{supcon}L_{supcon}+\lambda_{sty}L_{sty}+\lambda_{con}L_{con}+\lambda_{lpips}L_{lpips}$, where $\lambda_{clip},\lambda_{clip_f},\lambda_{supcon},\lambda_{sty},\lambda_{con}$, and $\lambda_{lpips}$ are coefficients to balance these loss components. 

%% file: Section/05_Experiments.tex
\section{Experiments}

\subsection{Implementing Details}
% \noindent $\bullet$  \textbf{Training strategy.} To achieve better visual quality, we use the VGG-19~\cite{VGG} pre-trained on the ImageNet dataset as the Image Encoder and fix its parameters. Then, we train \net in two stages: (1) We built an auto-encoder structure with the Image Encoder and Image Decoder, and train the Image Decoder for image reconstruction. (2) We add the Style Fusion module to train the whole network for style transfer.

\noindent $\bullet$ \textbf{Datasets.} We use the images from MS-COCO~\cite{COCO} (about 118k images) for the image reconstruction task in the first training stage. In the second training stage, we train \net with MS-COCO~\cite{COCO} as our content image set and WikiArt~\cite{WikiArt} (about 81k images) as the style image set. In the training phase, we load the images with the size of $512\times512$ and randomly crop them as training patches of size $256\times256$. As data augmentation, we randomly flip the content and style images. For inference, our \net can handle the images with any resolution. In this Section, we use the images with $512\times512$ resolution for a fair comparison.

\noindent $\bullet$ \textbf{Parameter setting.} We train  \net using Adam optimizer with the learning rate of $1\times10^{-4}$. The batch size is set to 30 and \net is trained for 100k iterations (about 8 hours) on a PC with one NVIDIA V100 GPU using PyTorch deep learning platform. The weighting factors in the total loss are defined empirically as: $\lambda_{clip}=1, \lambda_{lpips}=1, \lambda_{supcon}=2, \lambda_{sty}=50, \lambda_{con}=0.02$.

\noindent $\bullet$  \textbf{Metrics and evaluation.} We use CLIP loss (text/image-image cosine similarity)~\cite{language_2} to measure the semantic similarity between the target texts/images and stylized images. For artist awareness, we follow AST~\cite{ast} to compute the style transfer deception rate, which is calculated as the fraction of generated images that are classified by the VGG-16 network as the artworks of an artist for which the stylization was produced. A higher value means closer to the artist's style. We also show image-driven experiments in the supplementary to demonstrate the versatility of our \net, which takes either text or image for stylization.

\subsection{Text-driven Artistic-aware Style Transfer}
\label{text-driven}
Feature extracted by the CLIP model has high-semantic discriminate power that can be used for similarity measurement. Therefore, we compute CLIP scores. First, we extract the feature embedding from the content, style, and transfer images by CLIP, then we compute content $s_{cont}$ and style $s_{style}$ scores, as defined by Equations~\ref{eqn:clip_cont}, for which the higher score means better performance ($\uparrow$ in the table). SSIM and VGG content loss are also used to compute content differences.
\begin{small}
\begin{equation} \tag*{(6)}
\label{eqn:clip_cont}
s_{cont}\left( I_c,I_{cs} \right) =\frac{E_{I}(I_c)\cdot E_{I}(I_{cs})}{\left\| E_{I}(I_c) \right\| \times \left\| E_{I}(I_{cs}) \right\|},\ \ s_{style}\left( t_s,I_{cs} \right) =\frac{E_{T}(t_s)\cdot E_{I}(I_{cs})}{\left\| E_{T}(t_s) \right\| \times \left\| E_{I}(I_{cs}) \right\|}
\end{equation}
\end{small}

To show the efficiency of our proposed approach, we compare our approach with AST~\cite{ast}, MGAD~\cite{mgad}, LDAST~\cite{language_1}, CLIPstyler~\cite{language_2} (CLIPstyler(fast) and CLIPstyler(opti)), DALL-E-2~\cite{dalle}, VQGAN-CLIP~\cite{language_7}, Stable Diffusion~\cite{ldm}\footnote{\url{https://beta.dreamstudio.ai/dream}, we utilized the official model that employs a content image and style prompt as input to apply the Stable unCLIP inference.}, TxST~\cite{txst}, DiffusionCLIP~\cite{diffusionclip} and DiffStyler~\cite{diffstyler} With default diffusion settings  (e.g., steps).

\input{Table/comparison_Artist}

\noindent \textbf{Quantitative Comparison.} 
Table~\ref{tab:artists} reports the results on $512\times512$ image stylization. We observe that CLIPstyler(fast) leads to the best content similarity score (0.736); however, the transferred images have poor artistic style performance. The results of AST have the second-best deception rate, but the similarity to the content image is only 0.538. CLIPstyler(opti), a slow but optimal version compared to CLIPstyler(fast), reaches good CLIP scores but achieves the worst deception rate. This is expected since CLIPstyler(opti) requires dedicated training for each individual content and style image. DiffusionCLIP~\cite{diffusionclip} leads to the best CLIP style score but with poor performance on content scores. TxST~\cite{txst,txst2} performs best on the deception rate but the CLIP style score is low. Overall, our \net provides a good balance between style and content, reaching the best performance on inference time (over 100$\times$ speedup) and second-best CLIP score and deception rate, which demonstrates its effectiveness. It is important to note that CLIPstyler(opti) MGAD, VQGAN-CLIP and DiffusionCLIP need very long training time for each artist; in contrast, \net does not need retraining, and therefore it is much faster than others. Figure~\ref{fig:deception_artist} shows the class-specific accuracy of the deception rate. We observe that using our \net (red bars) achieves better performance for all artists. 
% ==== Figure SOTAs Comparison
\begin{figure*}[t]
	\begin{center}
		\centerline{\includegraphics[width=1\columnwidth]{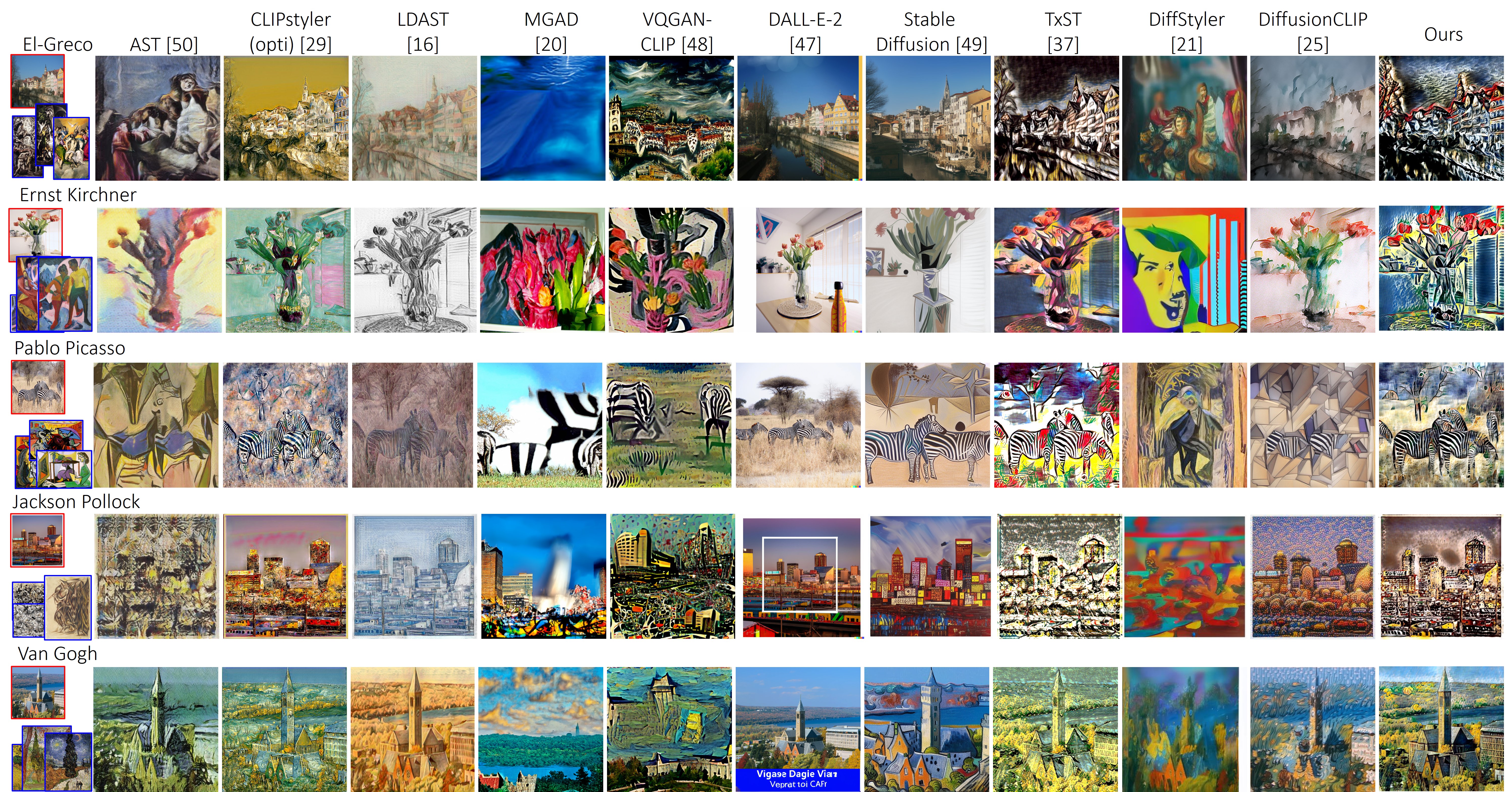}}
		\caption{\small
		\textbf{Compare among text-dirven style transfers.} We show five examples (content images in red boxes) using eight different methods. We use the names of painters as text prompts for style transfer. For reference, we also list representative paintings in blue boxes for comparison. The styles of artists are very abstract and subjective. We highly recommend readers check each artist online for better comparison. 
		}
		\label{fig:artist_compare_2}
	\end{center}
\end{figure*}

\noindent \textbf{User study.}
For user study, we invite users from different backgrounds, like art, design, literature, and science, to ensure the study is as fair as possible. The questionnaire compares the painting styles of eight artists from the WikiArt subset. For each artist, we collect the results from AST~\cite{ast}, CLIPstyler(opti)~\cite{language_2}, DALL-E-2~\cite{dalle}, VQGAN-CLIP~\cite{language_7}, Stable diffusion~\cite{ldm} and the proposed \net by using three different content images. In each question, the users were given three results from the same content image using the six methods. 
%Besides the artists' names, we provide users with three representative artworks from the artist as references. 
The users are requested to rank the style similarity to the artist, where 6 denotes the most similar, and 1 means the most different. We average the scores from all users as the Mean Opinion Score (MOS). The results are shown in Figure~\ref{fig:usr_study}. We observe that \net achieves the highest average MOS, 0.1$\sim$0.5 improvements compared to Clipstyler, Stable diffusion and TxST, which suggests that our results have the most similar painting style to the target artists \emph{at a perceptual level}. Quantitatively, we also utilize aesthetic scores~\cite{aesthetic} to rank all text-driven results to see which one is mostly consistent with human preferences. It was trained on over 238000 AI synthetic images with human rating scores. It can demonstrate that ours is more visually pleasant to humans.
\begin{table*}[t!]
\begin{minipage}[b]{0.65\textwidth}
\centering
\includegraphics[width=1\textwidth]{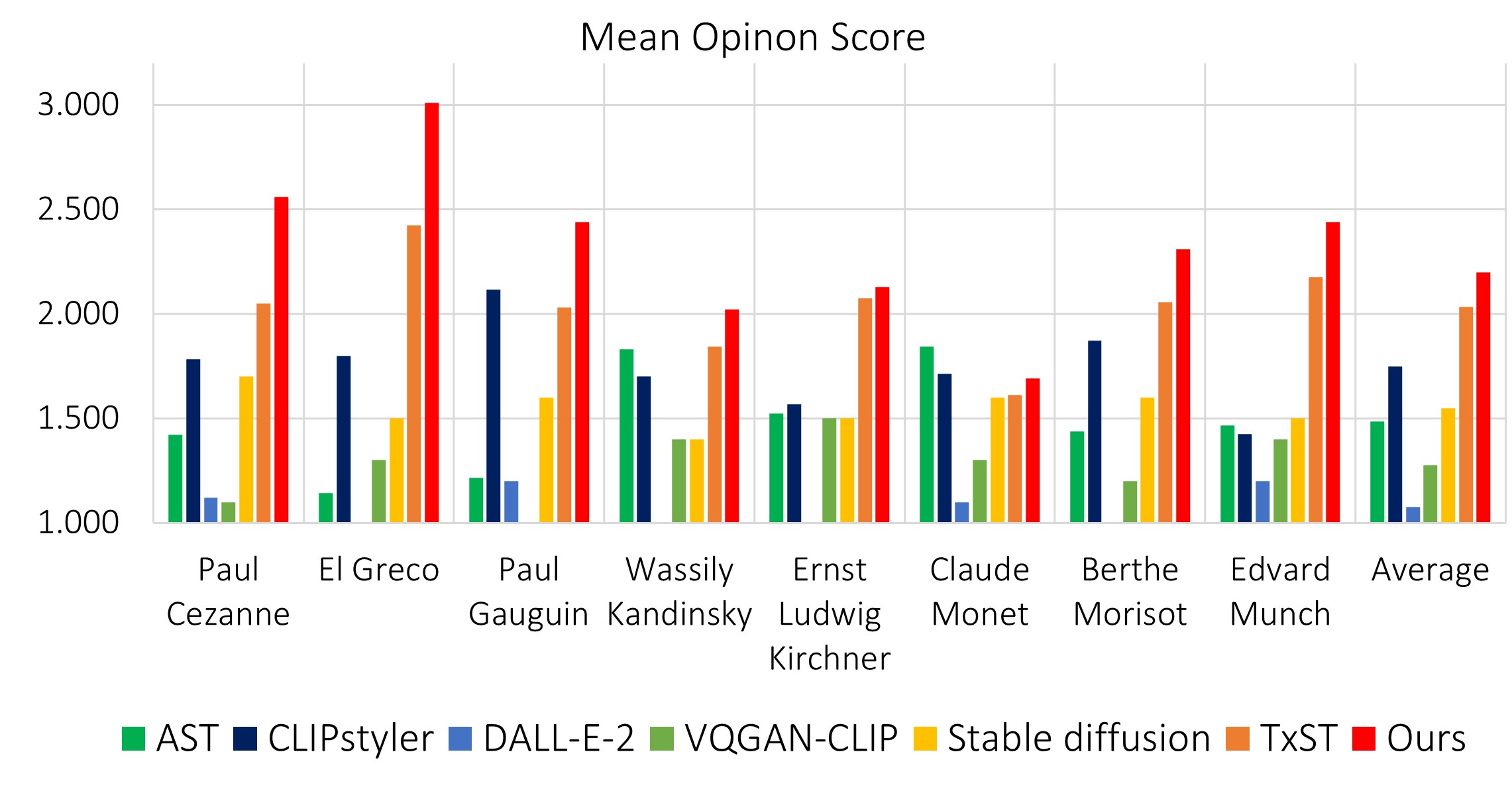}
\captionof{figure}{\small{\textbf{User study on text-driven style transfer.} We use the names of eight artists in WikiArt as text input for style transfer. We invite users to rank different approaches. The higher the Mean Opinion Score (MOS), the higher the style similarity to the target artists.}}
    \label{fig:usr_study}
\end{minipage}
\hfill
\begin{minipage}[b]{0.3\textwidth}
\centering
\footnotesize{
\resizebox{\textwidth}{!}{%
\begin{tabular}{l|c}
\hline
Method & Aes Score \\ \hline
AST~\cite{ast} & 4.976 \\
CLIPstyler~\cite{language_2} & 5.415 \\
DALL-E-2~\cite{dalle} & 4.052 \\
VQGAN-CLIP~\cite{language_7} & 4.360 \\
Stable Diffusion~\cite{ldm} & 5.268 \\
DiffusionCLIP~\cite{diffusionclip} & 4.368 \\
DiffStyler~\cite{diffstyler} & 3.723 \\
TxST~\cite{txst} &  5.436 \\
Our & {\color[HTML]{C00000}  {5.667}} \\ \hline
\end{tabular}%
}}
\captionof{table}{\textbf{Comparison on Aesthetic score.} The pretrained Aesthetic regression model~\cite{aesthetic} is used to rank different methods to see which one is most preferable to humans.}
\label{tab:aesthetic}
\end{minipage}
\vspace{2mm}
\end{table*}

\noindent \textbf{Qualitative Comparison.}
Figure~\ref{fig:artist_compare_2} shows the visual comparison of different methods for artist-aware style transfer. Note that AST does not require text input. For others, we use artists' names as texts to guide stylization. For reference, we also show three representative paintings in blue boxes. Our findings are summarized as: (1) our results show similar or better content preservation compared to AST and CLIPstyler. (2) \net can faithfully mimic the signature styles of specific artists, such as the color tone and temperature patterns from \textit{El-Greco, Van Gogh} and the distorted curves in \textit{El-Greco} and \textit{Van Gogh}. (3) VQGAN-CLIP, DiffStyler generates bizarre images resembling random generation. The official DALL-E-2 cannot perform style transfer but changes contents to some extent. Stable diffusion and DiffusionCLIP achieve visually pleasing results in \textit{P.\ Picasso} but it cannot learn the most representative style for other artists. 
% (4) \net can maximize the visual differences among different artistic styles (this is further discussed in the supplementary material). We also show all 13 artistic style transfer results on the same content image in Figure~\ref{fig:artist_compare}. We can see the visual diversity of different styles.

\input{Table/ablation_1}

\subsection{Ablation Study}
\label{sub:ablation}
\noindent $\bullet$  \textbf{Loss terms.} Here we ablate the losses \net is trained on and report the results in Table~\ref{tab:abla_loss}. \textit{Baseline} is the model that uses content $L_{\text{con}}$ and style $L_{\text{style}}$ losses. We observe that using either Directional CLIP loss ($L_{\text{CLIP}}$, 2nd row) or unsupervised contrastive loss ($L_{\text{unsup}}$, 3rd row) improves the CLIP style scores compared to the Baseline. This implies that these two losses can guide the stylization close to the target text description. By adding LPIPS loss ($L_{\text{lpips}}$), we can see from row 5 that it improves the VGG content loss and SSIM approximately by 3.9 and 0.09, respectively. The last row shows that our proposed supervised contrastive loss ($L_{\text{supcon}}$ can improve the CLIP style score by 0.05 with slight decreases in the VGG content loss and SSIM. 

\noindent $\bullet$  \textbf{Style Fusion Speedup.} We also compare the design of the style fusion module that can learn the cross-modal correlations between texts and images. From Table~\ref{tab:abla_mamba}, we compare ours (last row) with others, including the vanilla Attention~\cite{sanet} module and AdaIN~\cite{adain} structure. We can see that using our proposed style fusion module can speed up the training and testing by $1.8\times$ and $20\times$ with 59\% fewer parameters.

\subsection{Video style transfer and failure analysis}
\label{sub:video_st}
To demonstrate the style consistency of our method, one video sequence is applied to different style transfer approaches. To measure the motion changes, we use the open package Pyflow \footnote{\url{https://github.com/pathak22/pyflow}} and visualize the motion magnitude and direction. Figure~\ref{fig:video_st} demonstrates the results. We observe that using our method can obtain similar optical flows as the original video, which (1) firstly reveals its consistency in stylization, i.e., frames from the same video can remain the same artist's style and (2) secondly, highlights its content preservation, i.e., the motion changes do not affect much the stylization quality. On the other hand, other methods do not learn the correct style (\textit{Albrecht Durer} is famous for his woodcut print, for which other approaches fail to grasp this style) and they also fail to provide consistent stylization.
\begin{table*}[t!]
\begin{minipage}[b]{0.48\textwidth}
\centering
\includegraphics[width=1\textwidth]{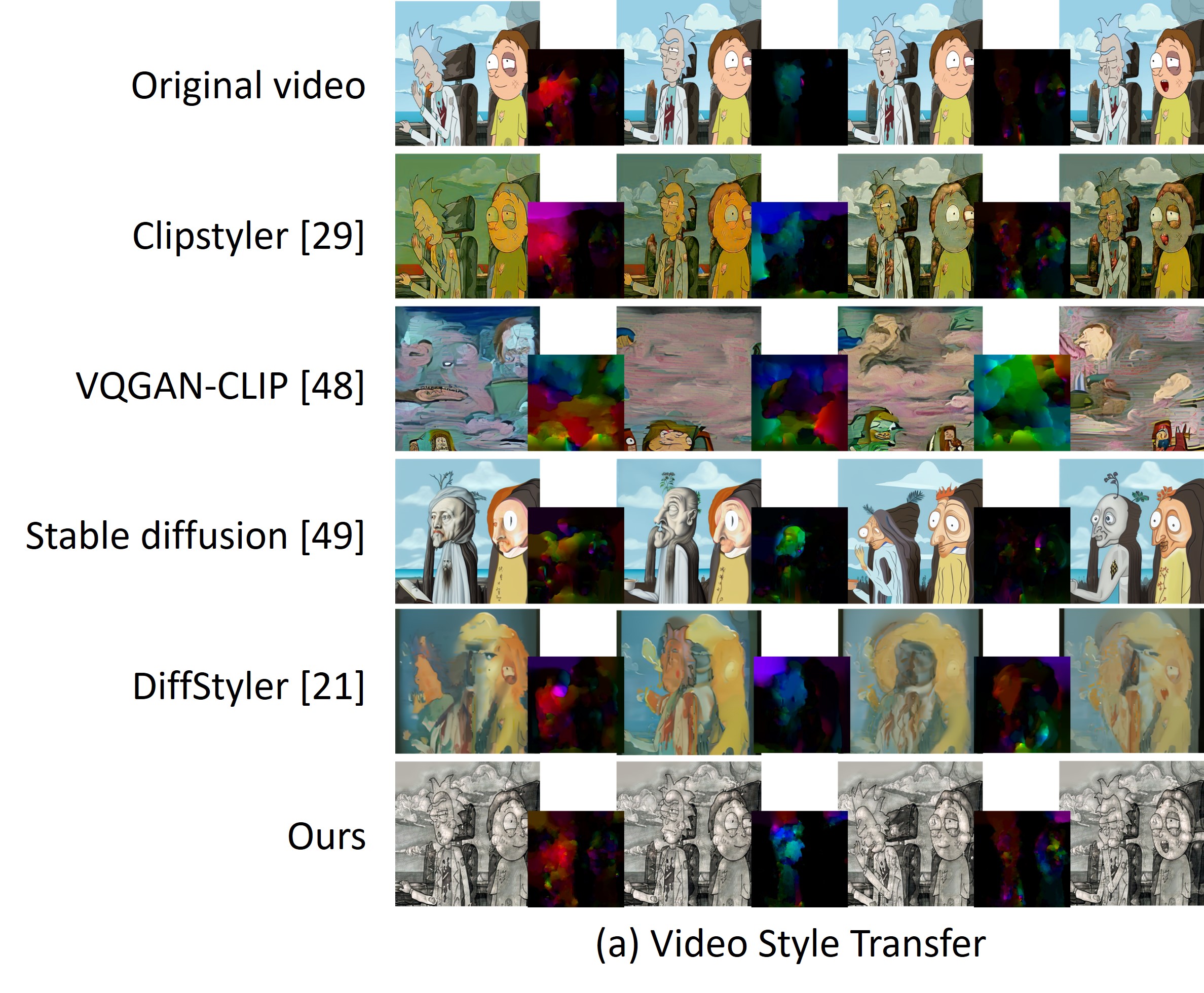}
    \captionof{figure}{\small{\textbf{Video-based text-driven style transfer.} Four neighborhood frames and their optical flows are shown. We use \textit{Albrecht Durer} as the text prompt to apply five different style transfer.}}
    \label{fig:video_st}
\end{minipage}
\hfill
\begin{minipage}[b]{0.48\textwidth}
\centering
\includegraphics[width=1\textwidth]{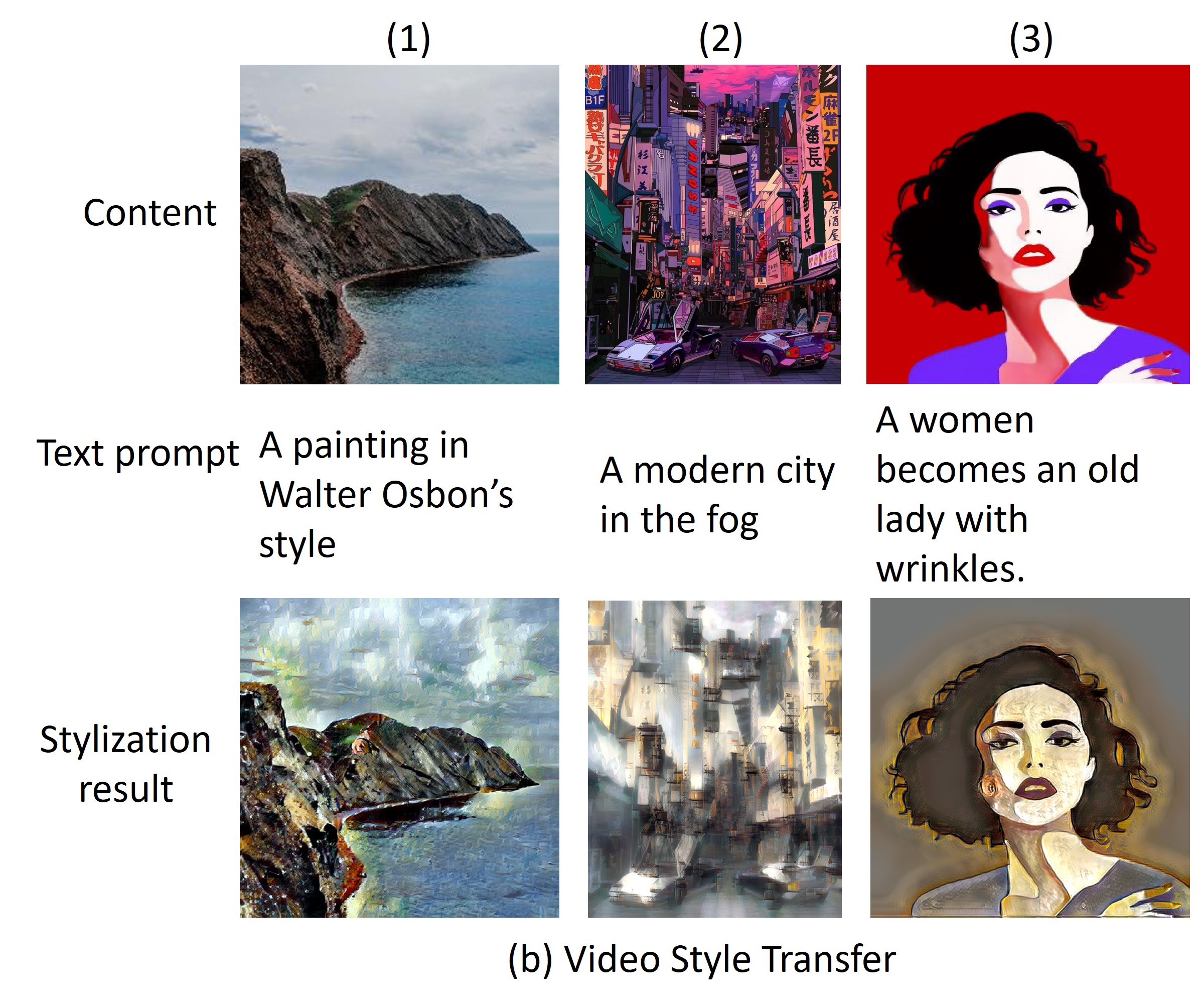}
\captionsetup{skip=0cm, width=0.99\textwidth}
    \captionof{figure}{\small{\textbf{Failure cases of text-driven style transfer.} Three failure cases are shown here, including unbalanced training, insensitive to general texts, and text prompt misunderstanding.}}
    \label{fig:failure}
\end{minipage}
\vspace{2mm}
\end{table*}

\noindent Overall, our proposed \net can perform well for text-guided style transfer. However, (1) for artistic style transfer, it performs poorly for artists with few examples in WikiArt~\cite{ast}. In Figure~\ref{fig:failure} (a), WikiArt only contains 20 paintings of \textit{Walter Osbon}, a painter of the post-impressionism movement. We observe that \net does not predict the most distinct style for stylization. (2) When the content image contains rich information, \net performs poorly for content preservation. In Figure~\ref{fig:failure} (b), though \net successfully adds fog to the image and blurs some contents, it also changes the color tone and loses some details. (3) \net may ignore texts with detailed descriptions. In Figure~\ref{fig:failure} (c), it misinterprets the ``old lady'' as an old picture and does not add wrinkles to the woman's face. All failure cases can be resolved by following the same pipeline: (i) training the model on a large-scale dataset, (ii) increasing the length of text prompts with more details, and (iii) computing losses for multiscale content preservation; this is, however, out of the scope of this work and we leave it for future work.

%% file: Table/comparison_Artist.tex
\begin{table*}[ht!]
\begin{minipage}[b]{0.48\textwidth}
\resizebox{\linewidth}{!}{
\begin{tabular}{l|ccccc}
\toprule
                                  & \multicolumn{2}{c}{Clip Scores}                                                  &  & Deception  & Running         \\
\multirow{-2}{*}{Method} & Content$\uparrow$             & Style$\uparrow$               & \multirow{-2}{*}{SSIM$\uparrow$}                 & Rate$\uparrow$    & %\multirow{-2}{*}
{time (s)}            \\ \hline
AST\cite{ast}                               & 0.538                        & 0.269                        & 0.235                        & 0.664 & 1.3 \\
CLIPstyler(fast)\cite{language_2}                  & {\color[HTML]{C00000} 0.736} & 0.254                        & 0.334                        & 0.469        & 0.9                \\
CLIPstyler(opti)\cite{language_2}                   & 0.624                        & 0.306 & 0.407 & 0.441    & 220                    \\
LDAST\cite{language_1}                   & 0.669                        & 0.207 & 0.255 & 0.435       & 1.6                 \\
MGAD\cite{mgad}                   & 0.397                        & 0.203 & 0.338 & 0.339          & 604              \\
VQGAN-CLIP\cite{language_7}                   & 0.557                        & 0.230 &  0.217 & 0.682        & 240                \\
DALL-E-2\cite{dalle}                   & 0.665                        & 0.228 & 0.358 & 0.425    & 34                    \\
Stable Diffusion\cite{ldm}                   & 0.542                        & 0.332 & 0.445 & 0.702   &   37                  \\
TxST\cite{txst} &
  {\color[HTML]{0000FF} 0.678} &
  0.313 &
  {\color[HTML]{0000FF} 0.487} &
  {\color[HTML]{C00000} 0.769} &
  0.7 \\
DiffusionCLIP~\cite{diffusionclip} &
  0.602 &
  {\color[HTML]{C00000} 0.443} &
  0.435 &
  0.669 &
  530 \\
DiffStyler~\cite{diffstyler} &
  0.540 &
  0.334 &
  0.301 &
  0.603 &
  6.5 \\
\midrule
\textbf{Ours} &
  0.667 &
  {\color[HTML]{0000FF} 0.402} &
  {\color[HTML]{C00000} 0.491} &
  {\color[HTML]{0000FF} 0.747} &
  {\color[HTML]{C00000} 0.03} \\ \bottomrule
\end{tabular}}
\captionof{table}{\small{\textbf{Text-driven style transfer comparison.} We use Clip style scores and the deception rate to measure the style similarity. Clip content score is used to measure content preservation.({\color[HTML]{C00000}  {Red}}: best and {\color[HTML]{0000FF}  {Blue}}: $2^{nd}$ best).}}
    \label{tab:artists}
\end{minipage}
\hfill
\begin{minipage}[b]{0.48\textwidth}
\centering
\includegraphics[width=1\textwidth]{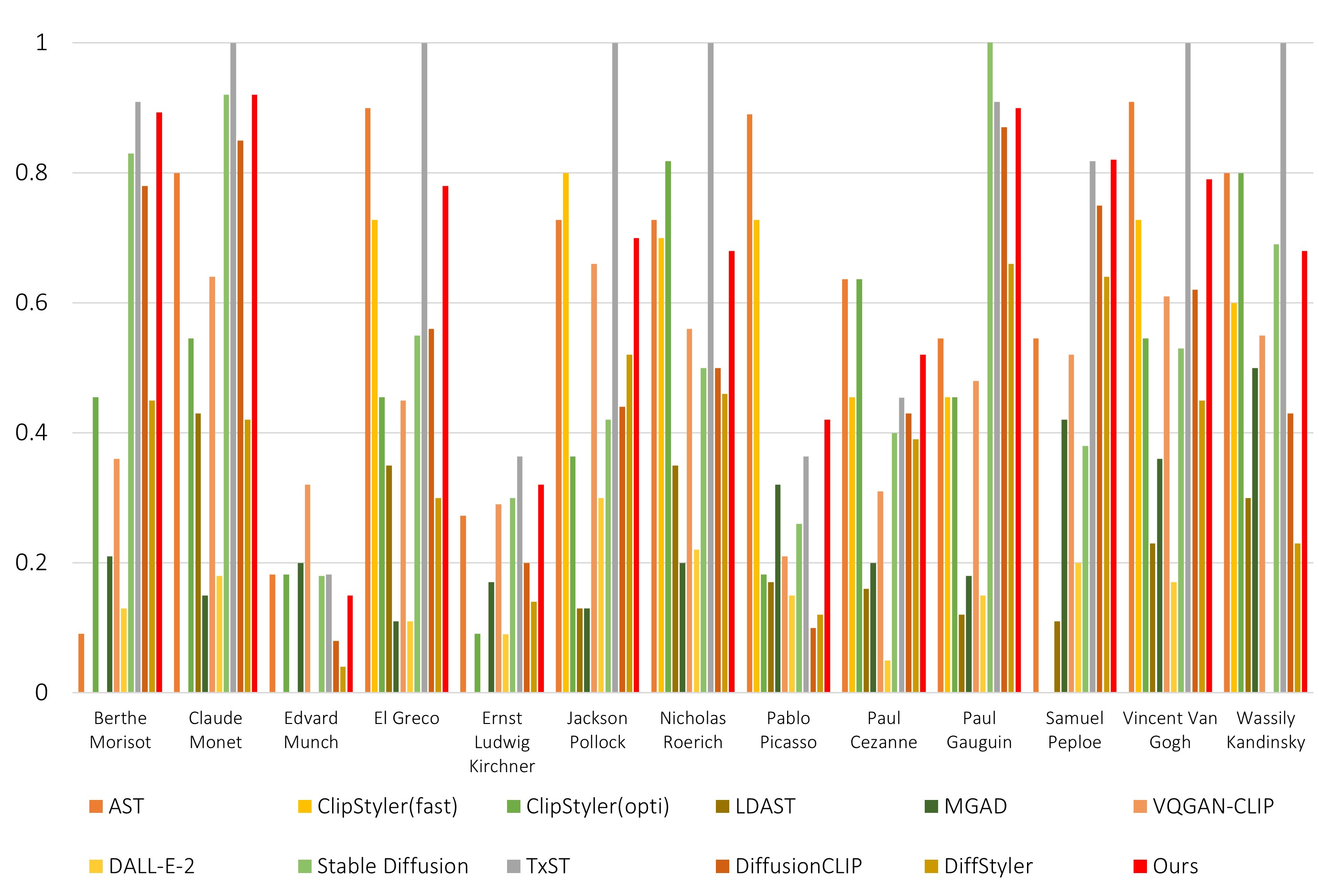}
\captionof{figure}{\small{\textbf{Deception rate of text-driven style transfer methods.} We show the deception rate of each artist from a WikiArt subset. Our method (red columns) achieves better performance than others. 
		}}
    \label{fig:deception_artist}
\end{minipage}
\vspace{2mm}
\end{table*}

%% file: Table/ablation_1.tex
\begin{table*}[t!]
\begin{minipage}[b]{0.465\textwidth}
\resizebox{\linewidth}{!}{
\begin{tabular}{l|ccc}
\toprule
\multicolumn{1}{c|}{Loss Terms} & \multicolumn{1}{c}{\begin{tabular}[c]{@{}c@{}}VGG $\downarrow$ \\ content\end{tabular}} & \multicolumn{1}{c}{\begin{tabular}[c]{@{}c@{}}CLIP$\uparrow$ \\ style score\end{tabular}} & \multicolumn{1}{c}{SSIM}$\uparrow$ \\ \midrule
Baseline ($L_{\text{con}}+L_{\text{style}}$) & 80.12 & 0.213 & 0.402 \\
Baseline+$L_{\text{clip}}$ & 86.40 & 0.355 & 0.383 \\
Baseline+$L_{\text{unsup}}$ & 82.05 & 0.218 & 0.400 \\
Baseline+$L_{\text{clip}}+L_{\text{unsup}}$ & 82.18 & 0.369 & 0.394 \\
Baseline+$L_{\text{clip}}+L_{\text{unsup}}+L_{\text{lpips}}$ & {\color[HTML]{C00000}78.26} & 0.354 & 0.481 \\ \midrule
Baseline+$L_{\text{clip}}+L_{\text{supcon}}+L_{\text{lpips}}$ & 78.56 & {\color[HTML]{C00000}0.402} & {\color[HTML]{C00000}0.479} \\ \bottomrule
\end{tabular}
}
\captionof{table}{\small{\textbf{Comparison on loss terms.} To show the effect of different loss terms, we compare the content preservation by VGG content loss and SSIM, and CLIP style score to measure style similarity.}}
    \label{tab:abla_loss}
\end{minipage}
\begin{minipage}[b]{0.49\textwidth}
\resizebox{\linewidth}{!}{
\begin{tabular}{l|ccc}
\toprule
Style Fusion module & \begin{tabular}[c]{@{}c@{}}Training $\downarrow$ \\ Time (hours) \end{tabular} & \begin{tabular}[c]{@{}c@{}}Inference $\downarrow$ \\ time (seconds)\end{tabular} & \begin{tabular}[c]{@{}c@{}}Number of\\ parameters\end{tabular} \\ \toprule
Attention+AdaIN & 8.7 (1$\times$) & 0.62 (1$\times$) & 2362112 \\
Attention+adaLN & 8.5 (1$\times$) & 0.62 (1$\times$) & 2393421 \\
Linear\_Attn+AdaIN & 6.5 (1.3$\times$) & 026 (1.2$\times$) & 1959408 \\
Linear\_Attn+adaLN & 5.7 (1.5$\times$) & 0.32 (1.2$\times$) & 1989237 \\
Mamba+AdaIN & 6.1 (1.4$\times$) & 0.03 (20$\times$) & {\color[HTML]{C00000}1399893} \\ \midrule
Mamba+adaLN & {\color[HTML]{C00000}4.9 (1.8$\times$)} & {\color[HTML]{C00000}0.03 (20$\times$)} & 1423104 \\ \bottomrule
\end{tabular}
}
\captionof{table}{\small{\textbf{Comparison on style fusion modules.} The training time starts from the same initialization, to
reach the target CLIP score of 0.4. Both training and inference are tested on one V100 GPU.}}
    \label{tab:abla_mamba}
\end{minipage}
\vspace{2mm}
\end{table*}

%% file: Section/06_Conclusion.tex
\section{Conclusion} 

In this paper, we proposed a text-driven approach for artist-aware style transfer, coined CLAST. To extract style descriptions effectively from the CLIP-based image-text space, CLAST leverages a supervised contrastive similarity training strategy and a new adaLN-SSM based module for style fusion. This approach explores the inherent relationship between texts and images by clustering artistic styles into different groups, eliminating the need for extensive data collection and online training. Extensive results show that CLAST achieves perceptually pleasing arbitrary stylization in real time, revealing its ability to extract critical representations from the CLIP space and produce aesthetics close to the artists' works. CLAST also points to a new direction for text-driven style transfer. Future work includes combining images, texts, and other cues to deliver a more flexible user-guided style transfer.